\documentclass[twoside,11pt]{article}
\pdfoutput=1

\usepackage{jmlr2e}
\usepackage{caption}
\usepackage{subcaption}
\usepackage{amsmath}
\usepackage{cleveref}

\begin{document}

\title{A Review of Scene Representations for Robot Manipulators}

\author{\name Carter Sifferman \email sifferman@wisc.edu \\
       \addr Department of Computer Sciences\\
       University of Wisconsin - Madison}

\maketitle


\section{Introduction}
For a robot to act intelligently, it needs to sense the world around it. Increasingly, robots build an internal representation of the world from sensor readings. This representation can then be used to inform downstream tasks, such as manipulation, collision avoidance, or human interaction. In practice, scene representations vary widely depending on the \textit{type of robot}, the \textit{sensing modality}, and the \textit{task} that the robot is designed to do. This review provides an overview of the scene representations used for robot manipulators (robot arms). We focus primarily on representations which are built from real world sensing and are used to inform some downstream robotics task.

Building an intermediate scene representation is not necessary for a robotics system. It is completely possible for a robotics system to act directly on sensor data (e.g. predict appropriate grasps directly from RGB images), and we will look at many such systems within this review. However, we argue that intermediate scene representations are beneficial for robot manipulators as they:
\begin{itemize}
    \item act as \textbf{spatial memory}
    \item are \textbf{efficient storage} of past memories
    \item allow \textbf{long-horizon planning}
    \item can act as \textbf{regularization} and encode \textbf{spatial priors} for learning systems
\end{itemize}

\begin{figure}
    \centering
    \includegraphics[width=\textwidth]{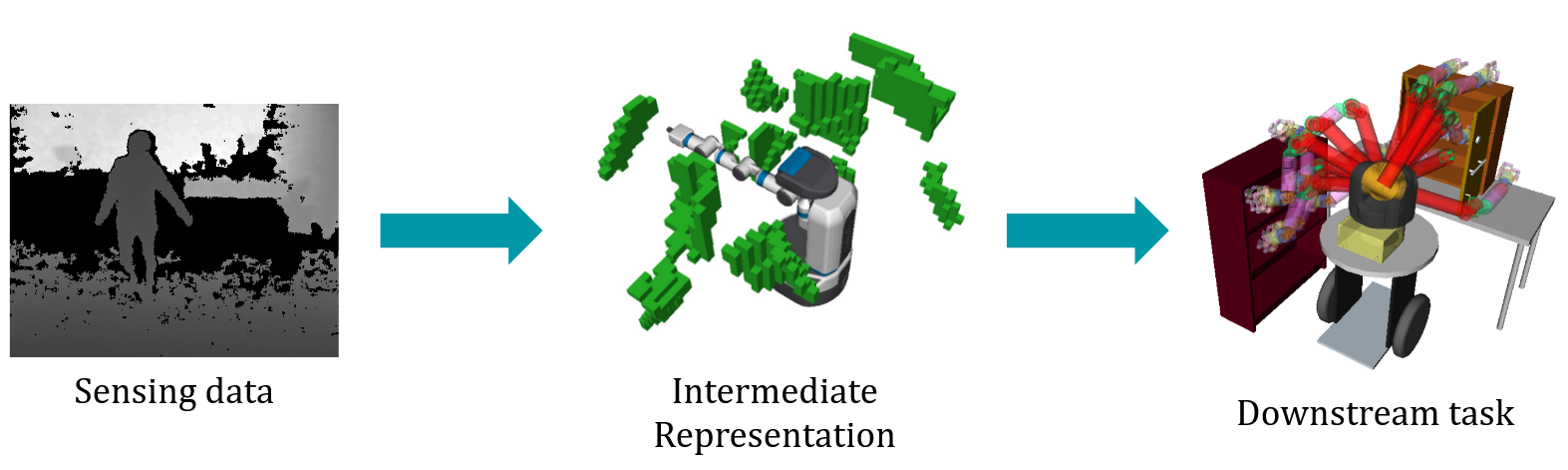}
    \caption{This review focuses on robotics applications which build some intermediate scene representation between sensor data and a downstream robotics task. Image sources: \cite{Moore2019, Kang2022RCIK, ratliff2009chomp}}
    \label{fig:diagram1}
\end{figure}

In this review, we organize scene representations into three categories depending on the task that the representation supports. These categories make up the sections of our review: collision avoidance (\cref{sec:collision_avoidance}), manipulation (\cref{sec:manipulation}), and teleoperation (\cref{sec:teleop}). Within each section, we provide a review of the existing literature, summarize the challenges in the area, and consider directions for future research. In \cref{sec:future_research}, we look at scene representations for manipulators as a whole, and consider directions for future research which cut across our three categories.

\subsection{Literature Survey Process}
Our literature survey consisted of two phases. In phase 1, we performed a broad search of existing reviews in order to gain context and understand the broader landscape of robotics. To find these reviews, we use Google Scholar search with the ``Review Articles'' filter enabled. The result of this search is 13 reviews spanning a broad spectrum:  manipulation and grasping \cite{landsiedel2017review, Kroemer2022, Cong2021, Billard2019, Cui2021, Guerin2020, mason2018toward}, SLAM \cite{Rosen2021}, human-robot interaction \cite{Fan2022, ajoudani2018progress}, teleoperation \cite{Wonsick2020}, motion planning \cite{Hwang1992, oleynikova2016signed} and inverse kinematics \cite{Aristidou2018}. Knowledge gained from these reviews was used to determine our taxonomy of scene representations, and the scope of this review.

The goal of phase 2 of our survey process was to find directly relevant papers which employ a scene representation for robot manipulators. As the papers that we search span a wide range of topics, keyword searches did not prove effective. Instead, we ``snowballed'' through references, beginning at seed papers which were found through keyword search, reviews, or consulting with colleagues. These seeds are shown in \cref{tab:searches}. We snowballed through references both through traditional reference chasing, and using Google Scholar's ``Cited By'' page to find papers published after the seed paper. We also use the Abstract Viewer Project\footnote{https://pages.graphics.cs.wisc.edu/AbstractsViewer/}, a system for finding related papers developed as a research project at UW-Madison. Abstract Viewer does not use citations to find related papers, instead relying on analysis of textual content. This proved helpful for finding still-related papers when references ran dry. In total, 69 papers were collected in phase 2.

This review is not meant to act as a comprehensive survey of any one subject. Instead, we hope to give a broad overview of the field and provide jumping-off points for further reading. Articles were not selected to be a representative sample of the entire field; papers which use an identical scene representation to existing work may be excluded while those with a unique scene representation are generally included.

\begin{table}[]
    \centering
    \begin{tabular}{l|c|c}
         \textbf{Category}  & \textbf{Snowball Seeds} & \# \textbf{Papers Found} \\
         \hline
         Collision avoidance  & \cite{Kang2022RCIK, Rakita2021CollisionIK, ratliff2009chomp} & 26 \\ 
         Manipulation & \cite{Simeonov2021, xu2020learning, Saxena2006} & 28 \\ 
         Teleoperation & \cite{Wei2021, livatino2021intuitive, Wonsick2020} & 16 
    \end{tabular}
    \caption{The three sections that this review is organized into, and the ``snowball seeds'' which began the literature review process.}
    \label{tab:searches}
\end{table}

\section{Collision Avoidance}
\label{sec:collision_avoidance}

A central challenge in robotics is being able to avoid collisions, which can potentially be very costly with a powerful, fast, and expensive robot. Some approaches for collision avoidance respond directly to measurements from robot-mounted sensors, without building any intermediate representation \cite{Kroger2010, Avanzini2014, Tsuji2019}. A downside of this approach is that it has no memory: the robot can only act based on what it currently observes, and cannot plan based on its previous observations. As a result, these approaches are overly cautious and perform poorly in challenging conditions.

Modern approaches for collision avoidance can be broadly grouped into two categories: motion planning \cite{lavalle_2006}, in which the start and goal position of the robot end effector are known, and the goal is to find a viable, collision-free path between the two, and inverse kinematics, in which the end effector position is known, and the goal is to find a viable joint configuration of the robot which matches that end effector position. This distinction is unimportant for this review, as both of these approaches have the same requirements of their scene representations: fast collision checks and (sometimes) fast calculation of the distance to the nearest obstacle or direction to the nearest obstacle. As a result, scene representations for collision avoidance are similar whether the problem is formulated as inverse kinematics or motion planning.

\textit{Potential Fields--} Early approaches to collision avoidance in a motion planning context represented the environment with a potential field, first proposed in \cite{Khatib1985}. This potential field can be evaluated at any point to yield a scalar ``potential'' value. This value is determined by both the scene geometry (which have high potential around them) and the desired location (which has a low potential). In order to move through space, the robot simply follows the negative gradient of this potential field. The potential field was heavily utilized in early motion planning work \cite{Hogan1984, miyazaki1984sensory, pavlov1984method}. While effective for the time, potential fields suffer from a few problems: the potential function is prone to local minima, and can be very difficult and computationally intensive to construct \cite{Hwang1992}. Additionally, it is impractical to construct potential fields in real-time from sensor measurements, both because they are slow to construct, and because they require scene geometry to be described in a friendly closed-form, which sensors cannot provide natively. Later work \cite{Ge2000} improved on the local minimum problem by taking into account the relative starting position as well as scene geometry during potential field construction, but nonetheless potential fields have fallen out of favor in collision avoidance applications since the early 2000s.

\textit{Signed Distance Fields--} A signed distance field is a mapping between 3D points in space $\textbf{x}$ and the scalar distance $d$ to the nearest obstacle:
\begin{equation*}
    SDF(\textbf{x}) = d
\end{equation*}
The SDF has the nice property that taking $-\nabla SDF(\textbf{x})$ yields a vector pointing towards the nearest obstacle. This representation has been used in computer graphics since at least 1998 \cite{Frisken2000, Gibson1998}, and became popular for robot collision avoidance with the introduction of the highly influential CHOMP motion planner in 2009 \cite{ratliff2009chomp}. CHOMP uses the signed distance field, along with pre-computed gradients to perform optimization over the robot configuration. Subsequently, the popular STOMP \cite{kalakrishnan2011stomp}, TrajOpt \cite{schulman2013finding}, and ITOMP \cite{Park2012} motion planners also used a signed distance field to represent their environment, and used gradients in a similar way. In each of these works, the signed distance field is computed at fixed points on a regular voxel grid as a pre-processing step. To do such a computation, a precise model of the underlying geometry is needed, typically in the form of a mesh. Similarly to potential fields, computing a signed distance field is computationally expensive, and generating it from noisy sensor data is difficult. In practice, collision avoidance approaches which use an SDF are constrained to simulations, where the SDF can be pre-computed, or static environments in the real world. Regardless, SDFs are the most popular scene representation for collision avoidance, largely because they are supported by popular and effective motion planners. There exist libraries such as VoxBlox \cite{Oleynikova2017voxblox} and FIESTA \cite{Han2019FIESTA} which efficiently compute and store discretely sampled SDFs for this purpose.


\textit{Collections of Primitives--} A less common method for representing geometry for collision avoidance is with a collection of primitive shapes, such as spheres, cylinders, and cubes. These primitives are usually stored parametrically, so that collision checking can be done quickly and the objects represented natively in optimization solvers. Toussaint et. al. \cite{Toussaint2015LogicGeometricPA} proposed Logic-Geometric Programming, in which the scene is composed entirely of parametric cylinders, blocks, and planes. This paper has been influential for its elegant optimization-centered formulation, but the scene representation used within has not been heavily utilized; it serves more as a demonstration of the approach's capability. Similarly, Gaertner et. al. \cite{Gaertner2021} considers collision avoidance with humanoid robots, and uses a collection of primitives to represent dynamic scenes. Similarly to Toussaint, the collection of primitives is used primarily to demonstrate the capabilities of the system. In contrast, Zimmerman et. al. \cite{Zimmermann2022} proposes a method for using collections of primitives in gradient-based optimization methods such as TrajOpt \cite{schulman2013finding}. They provide a unified method for dealing with many types of primitives, and a method for taking the derivative of the distance to the nearest primitive, making a collection of primitives a drop-in replacement for SDFs. However, this approach has not seen widespread adoption.


\textit{Collections of Convex Hulls--} A collection of convex hulls is another less common way to represent scene geometry for collision avoidance. Similarly to primitive shapes, convex hulls allow for fast collision checking and natural representation in optimization solvers. Convex hulls have the additional benefit that any shape can be broken down to a collection of convex hulls via an algorithm like QuickHull \cite{barber1996QuickHull}. Schulman et. al. \cite{Schulman2014} uses a set of convex hulls to represent a scene, and outperforms the SDF-based motion planners of the time like CHOMP \cite{ratliff2009chomp} and STOMP \cite{kalakrishnan2011stomp}. CollisionIK \cite{Rakita2021CollisionIK} introduces an optimization-based method for \textit{inverse kinematics}, which is able to operate in real-time (e.g. for mimicry control) and avoid collisions with dynamic obstacles. CollisionIK mentions that point cloud objects could be broken down into convex hulls in real time, but does not demonstrate such a process. To our knowledge, no existing approach constructs convex hulls in real time from sensor data.


\textit{Learned Representations--} Within the last year, one approach has emerged which uses a learned environment representation to enable real-time collision avoidance with dynamic obstacles and real-world sensing. This paper is somewhat influenced by the growing literature around learned representations in computer vision \cite{mildenhall2021nerf}, graphics \cite{lombardi2019neural}, and SLAM \cite{Sucar2021iMAP, Zhu2022NICE-SLAM}. RCIK \cite{Kang2022RCIK} proposes a collision cost prediction network, a neural network which takes as input features extracted from an occupancy grid, as well as a 3D point in that grid; from this input the network predicts the collision cost, which is an approximation of the SDF evaluated at that 3D point. The occupancy grid can be generated in real time with one or more depth cameras. The network is trained on one million simulated examples of random environments and joint configurations. While this method does not have the collision avoidance guarantees provided in theory by other methods, it is the first method to perform collision-free inverse kinematics in real time with real sensing. This same approach was later evaluated under real-time control \cite{kang2022}.

\begin{figure}
    \centering
    \includegraphics[width=0.75\textwidth]{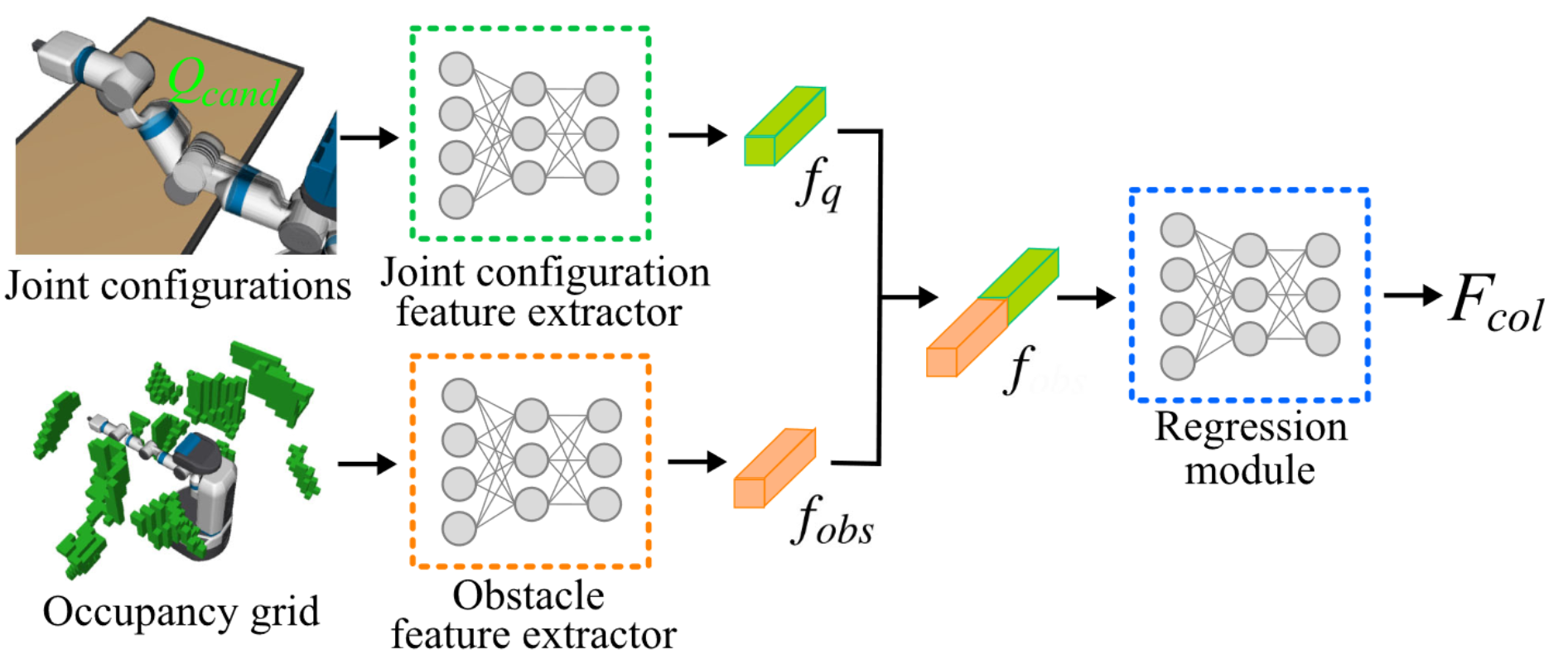}
    \caption{The collision cost prediction network of RCIK \cite{Kang2022RCIK} is trained on simulated data, and outputs a collsion cost $F_{col}$ which approximates the signed distance function to enable real-world real-time collision avoidance.}
    \label{fig:my_label}
\end{figure}


\subsection{Future Directions in Collision Avoidance}
\textbf{Real-time generation of signed distance fields}. Signed distance fields have proven highly effective for enabling collision avoidance. However, SDFs are very costly to generate, and require a very accurate representation of the underlying environment. Because of this, the vast majority of work on collision avoidance with robot manipulators does so only in simulation, or in manually recreated static environments. A pressing challenge is finding ways to bring these methods to the real world by constructing an SDF in real time. Recent work on neural representations has enabled real-time construction of a neural SDF from depth imagery called iSDF \cite{Ortiz2022iSDF}. Similarly to RCIK \cite{Kang2022RCIK}, the SDF produced by iSDF is only an approximation, however it is generated over time from multiple sensor observations, and does not rely on simulated pre-training. Adapting a similar approach for static or manipulator-mounted depth cameras could enable real-world operation of the many collision avoidance approaches which rely on SDFs. 

\textbf{Moving beyond signed distance fields}. Signed distance fields alone are great for collision avoidance, but offer some limitations. For example, not all collisions are equally costly. Colliding with a pillow might be admissible if it means avoiding collision with a human. Future representations, and algorithms which act on them, could store semantic information along with scene geometry, to enable such decisions to be made. Learning such semantic scene properties may be possible via extensive pre-training, or via interaction.

\section{Manipulation}
\label{sec:manipulation}
Arguably the most important task for a robot \textit{manipulator} is to \textit{manipulate} things. Manipulation can mean grasping with a simple one degree-of-freedom gripper, articulated grasping, or simple pushing and nudging of objects with any part of the robot. Robot manipulation is a vast field, with approaches specialized for dealing with many specific challenges. To keep the scope of this review reasonable, we focus on scene representations which are used for:
\begin{itemize}
    \item Basic grasping with a 1DoF gripper
    \item Generalizable grasping
    \item Articulated grasping
    \item Predicting scene flow
\end{itemize}

\textit{Direct Action on Images--}
While this review focuses on intermediate scene representations, it is clear that, for the purposes of robot manipulation, an intermediate representation is not necessary. A seminal work in this area was Saxena et. al. \cite{Saxena2006} in 2006, which was the first work to directly predict grasping points from an image. They use a neural network to, given an RGB image of an object to be grasped, predict pixels in the image at which the object is most suitable for grasping. To train their neural network, they use supervised learning on a large fully synthetic dataset. Two years later, the same authors improved on the approach by using RGB-D imagery as input to the network \cite{Saxena2008LearningGS}. Lenz et. al. \cite{lenz2015deep} improved upon previous work by aiming to predict the single best grasp, rather than listing many viable ones, and predicting the orientation and extent of the grasp along with the location. Later, the ``DexNet'' series of papers offered iterative improvements by improving training data and tweaking the neural network outputs, as well as considering alternative grip types such as articulated hands and suction cups \cite{Mahler2017DexNet2D, Mahler2017DexNet3, Mahler2019DexNet4}. Other work aims to grasp objects given some semantic label, e.g. ``coffee mug'' or ``plate'', this is sometimes called semantic grasping. Jang et. al. \cite{Jang2017} proposes a two-stream approach to semantic grasping, in which one stream identifies objects while another finds suitable grasps. Schwarz et. al. \cite{schwarz2018fast} demonstrates a semantic grasping pipeline which uses a suction cup gripper and works by simply segmenting out objects and finding their center of mass. This approach performed well at the highly competitive Amazon Picking Challenge\footnote{https://robohub.org/amazon-picking-challenge/}. These direct prediction approaches are highly effective for real world operation, however they are fairly limited in their potential. These approaches can only perform single-shot grasping, meaning they are unable to, for example, rotate an object, let go of it, and grab it again. They also have a limited ability to reason about novel shapes in 3D. Lastly, these approaches typically rely on some synthetic training data, which must be generated via other 3D-aware methods.

\begin{figure}
    \centering
    \includegraphics[width=0.5\textwidth]{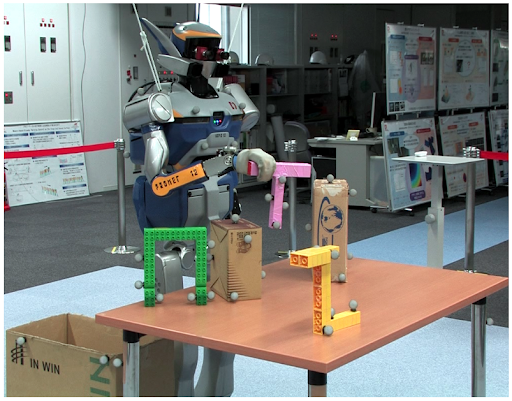}
    \caption{An early approach for real world grasping, Berenson et. al. \cite{berenson2007grasp}, relied on a motion capture system and pre-defined meshes to perform grasping in the real world.}
    \label{fig:Berenson}
\end{figure}
\begin{figure}
    \centering
    \includegraphics[width=\textwidth]{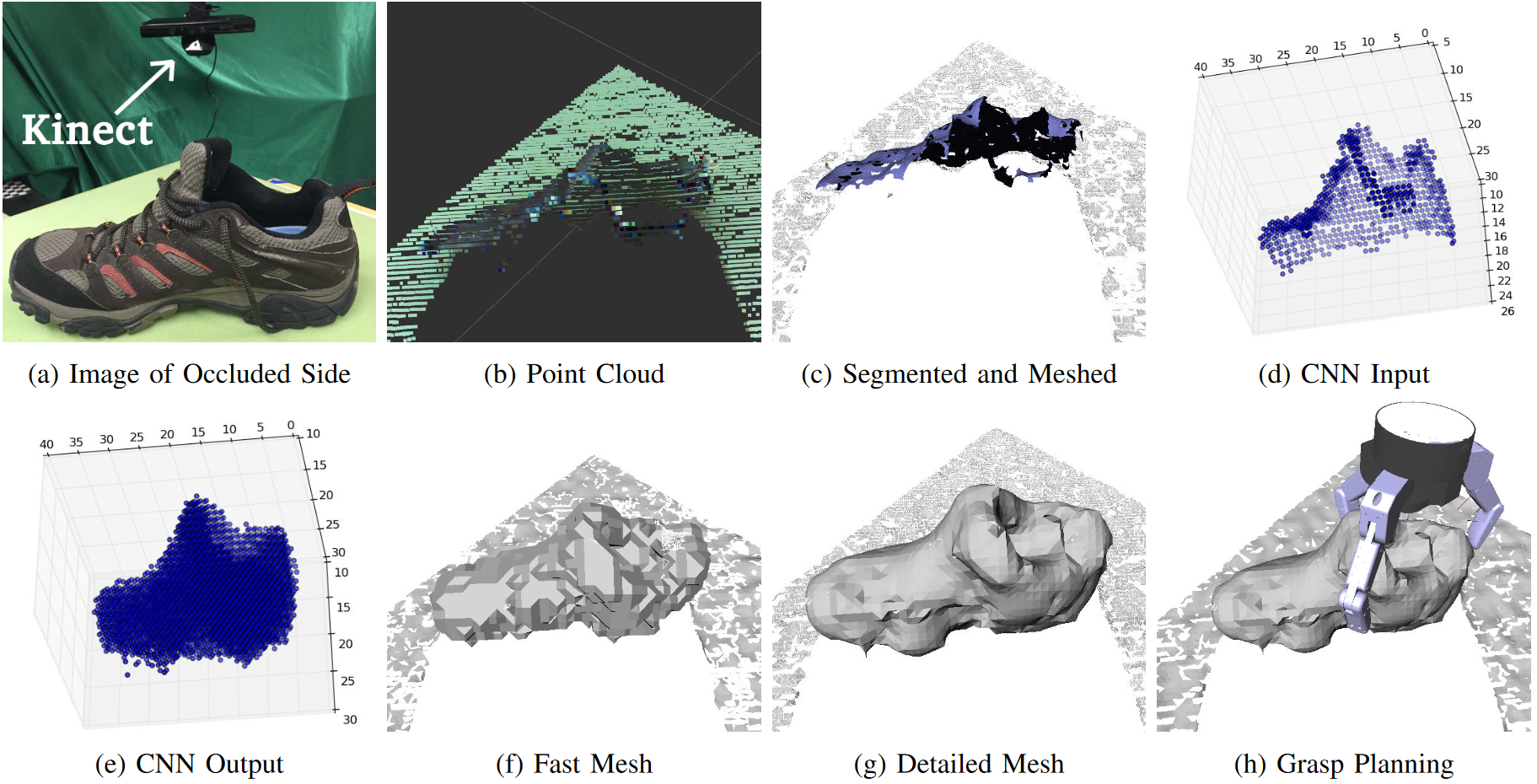}
    \caption{Varley et. al. \cite{Varley2016Shape} uses a neural network to enable shape completion on partial point cloud observations. The completed shapes can then be transformed to a mesh and used in any mesh-based grasp planner. Image from Varley et. al.}
    \label{fig:varley}
\end{figure}

\textit{Meshes--}
A number of works use a 3D triangular mesh to represent objects for manipulation. The problem of finding suitable grasps given a 3D mesh is a long-standing problem with active research \cite{dogar2012physics, Duenser2018, goldfeder2009columbia, Pokorny2013GraspMS, Weisz2012}. This paragraph focuses not on those algorithms, but on real-world systems for manipulation which represent objects as meshes. The first of such real-world systems was Berenson et. al. \cite{berenson2007grasp} in 2007. This work makes grasping possible in the real world by incorporating information about not only the object to be grasped, but also nearby obstacles, such as the table or other objects, into the grasp selection algorithm. In order to sense the positions of objects in real-world tests, this approach relies on motion capture markers being placed on each object, as shown in \cref{fig:Berenson}. Goldfeder et. al. \cite{Goldfeder2009} introduced a method for finding good grasps given a mesh, and tested their method by scanning real objects. This approach was somewhat effective, but the conversion from scan to mesh does not happen in real time. Collet et. al. \cite{collet2009object} circumvents the problem of real-time mesh construction by modeling each object as a primitive, and fitting that primitive to a point cloud in real time. Later work by the same authors \cite{Collet2011TheMF} extends this idea to arbitrary meshes by using a 6D pose recognition algorithm. Assuming that a mesh of the object is known, this approach enables prediction of the mesh's pose in real time. Papazov et. al. \cite{Papazov2012} takes a similar approach, but assumes that the object is represented with a set of points and surface normals, rather than an RGB image. Varley et. al. \cite{Varley2016Shape} removes the requirement of a pre-made mesh by teaching a neural network to complete a point cloud. From the completed point cloud, a mesh can be built and that mesh passed off to a mesh-based grasp planner in real time. The steps in Varley et. al.'s pipeline are shown in \cref{fig:varley}. In each of these approaches, we see that the limiting factor is not the grasp planning algorithms themselves, but the generation of meshes in real time.

\textit{Point Clouds--} Point clouds have been used as the basis for grasp planning in a similar manner to meshes. In constrast to meshes, point clouds are closer to the native output of commonly used depth cameras, making them more practical for real world construction. Approaches exist for grasp planning directly on point clouds \cite{Pas2017, ni2020pointnet++, duan2021robotics, ten2018using}, although less numerous than those for meshes. Florence et. al. \cite{Florence2018} (covered in more depth in the following paragraph) introduces a method for finding corresponding grasps between similar objects, and uses point clouds to perform the grasping in real-world examples. Simeonov et. al. \cite{Simeonov2020} considers the problem of manipulation planning, and is able to predict the movement of scene objects directly from their point clouds. This prediction is used to plan for manipulations which move the scene towards a goal state.


\textit{Voxel Grids--} Zhang et. al. \cite{Zhang2010GraspEW} constructs a gripper-sized voxel grid from a point cloud and a given gripper position, with each voxel encoding whether the space is occupied by a point in the point cloud. This voxel grid can then be compared (via nearest neighbors) to previously simulated voxel grids to determine whether it corresponds to a good gripper position.


\textit{Learned Representations--} Dense object nets \cite{Florence2018} are an approach for generalizable grasping. For a given input image, a neural network is trained to produce a dense pixel-level feature map which produces similar output feature vectors for semantically similar points in multiple images of similar objects. For example, for any two images of a mug, the goal is that pixels corresponding to the same point on the mug handle will have the same feature vector in the output representation. This representation is trained to be consistent across object instance, pose, and deformation, enabling generalizable grasping. They demonstrate that this approach is effective at generalizable grasping in the real world, using merged point cloud data from multiple depth cameras. Simeonov et. al. \cite{Simeonov2021} expands upon this approach by taking a 3D point as input to the neural network, rather than a 2D pixel position. The object properties are encoded within the weights of a multi-layer perceptron, similarly to NeRF \cite{mildenhall2021nerf}. They demonstrate that this approach can be used to perform pick-and-place tasks of novel objects of a given class given fewer than 10 demonstrations. Additionally, the architecture of their MLP ensures that the descriptor fields are SE(3)-equivariant, making them robust to arbitrary object poses. Van Der Merwe et. al. \cite{vandermerwe2020learning} uses a learned representation for articulated grasping. A neural network is trained to, given a point cloud and a 3D query point, approximate the signed distance function at that point. The latent space of this network is concatenated with information about desired grasp qualities and robot configuration, and fed into another network which predicts the success rate of the proposed grasp. Xu et. al. \cite{xu2020learning} builds a visual predictive model for robotics. Their goal is to, given some representation of the scene along with a robot action, predict the scene after the action has occurred. Rather than using a manually constructed scene representation, they allow the scene representation to be learned in an end-to-end manner, as a 128x128x48 feature grid with an 8-dimensional vector at each point in the grid. Using this representation, they are able to predict 3D scene flow and plan manipulation tasks.

\subsection{Future Directions in Manipulation}
\textbf{Performing 3D grasp planning based on real-world sensor data}. There is a disconnect between the way scenes are represented for grasp planning (primarily meshes) and the way that the most effective robotics systems, such as those used in the Amazon Picking Challenge, are operated (primarily direct prediction). An important direction for future work is bridging this gap. Direct prediction methods are severely limited in their spatial reasoning, while mesh-based methods are severely limited in their real-world operation. Point cloud manipulation and learned representations may bridge the gap, but do not currently offer the high grasping accuracy of mesh based approaches. 

\textbf{Representing complex object properties}. Any object may have many properties which are relevant to manipulation: its center of mass, deformation properties, or constraints on its motion (e.g. hold a mug full of coffee upright, pull a drawer straight out). Learning-based approaches, such as Dense PhysNet \cite{Xu2019} have shown promise towards being able to learn these properties autonomously. A needed direction for future research is determining ways to learn these properties, generalize them to novel objects, and store these properties alongside their geometric representations.

\section{Teleoperation}
\label{sec:teleop}
In a teleoperation scenario, a human controls a robot remotely, and relies on an intermediate interface, such as a monitor or VR headset, to understand the robot's surroundings. Much recent work on teleoperation places the human operator in virtual reality (VR); studies have shown increased task performance in virtual reality, due to the ease of controlling the user's view and all six degrees-of-freedom of the robot \cite{whitney2020comparing}. However, the scene representations used in these VR approaches can generally applied to any sort of teleoperation scenario.

\subsection{Human Control of Mobile Robots}
Mobile robots are much more likely than robot manipulators to venture into unknown environments. Because of this, the foundational work in representing scenes to remote operators has taken place in mobile robotics. For context, we offer a brief overview of such work here. Nielsen et. al. \cite{nielsen2007ecological} was an early attempt at incorporating data beyond RGB camera feeds into robot teleoperation. They create a dashboard which shows an RGB camera feed along with LiDaR scans and a rough map of the environment. While all of the information is presented to the user, it is not combined to create an intuitive display. Kelly et. al. \cite{kelly2011real} builds a 3D map of the environment by using LiDaR scans to create a 3D map, and coloring that map with data from RGB images. A CAD model of the mobile robot is placed in the environment. This allows rendering arbitrary viewpoints, such as an overview of the entire scene, an overhead view, or a third person view, as would be seen in a racing video game. Stotko et. al. \cite{stotko2019vr} builds a 3D mesh-based scene representation for a mobile manipulator, and displays the mesh to the user interactively in virtual reality. They find that users have fewer collisions, and report a greater level of immersion and awareness than with a 2D interface. Livantino et. al. \cite{livatino2021intuitive} augments robot-attached camera views with 3D data to overlay data such as desired path, destination, and label traversable terrain. For mobile robot teleoperation, the trend has been towards a free floating, user controllable view and natural display of information. These same goals apply to robot manipulator teleoperation.


\begin{figure}
     \centering
     \begin{subfigure}[b]{0.3\textwidth}
         \centering
         \includegraphics[width=\textwidth]{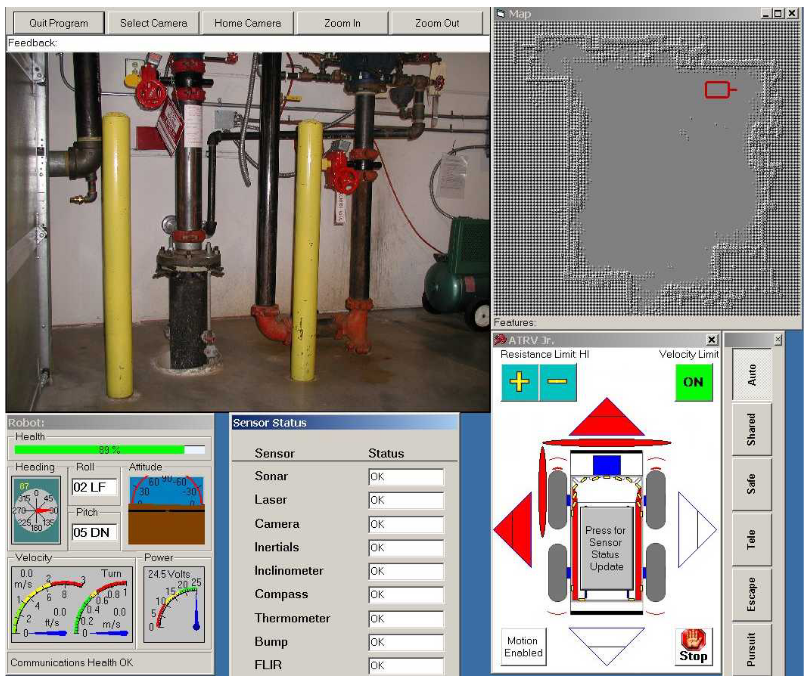}
         \caption{Dashboard-based interface from \cite{yanco2004beyond} (2004)}
         \label{fig:yanco2004}
     \end{subfigure}
     \hfill
     \begin{subfigure}[b]{0.3\textwidth}
         \centering
         \includegraphics[width=\textwidth]{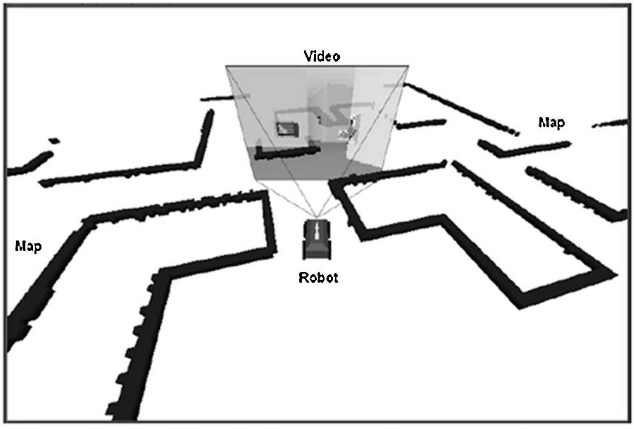}
         \caption{Augmented interface from \cite{nielsen2007ecological} (2007)}
         \label{fig:nielsen2007}
     \end{subfigure}
     \hfill
     \begin{subfigure}[b]{0.3\textwidth}
         \centering
         \includegraphics[width=\textwidth]{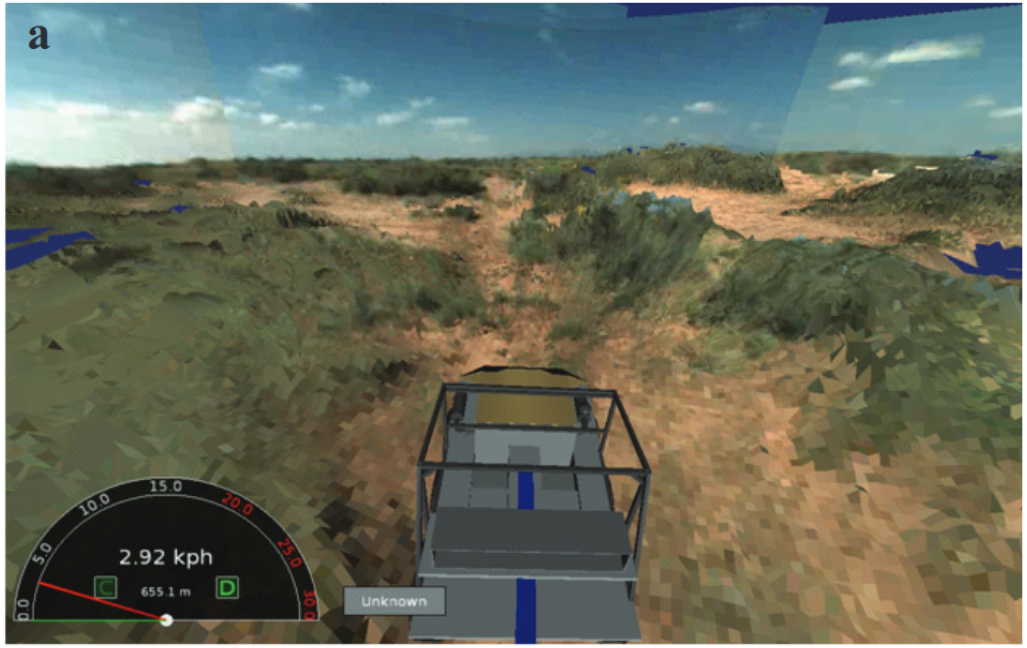}
         \caption{3D interface from \cite{kelly2011real} (2011)}
         \label{fig:kelly2011}
     \end{subfigure}
    \caption{A history of the typical information displays used in mobile robot teleoperation}
    \label{fig:mobile_robot_representations}
\end{figure}

\subsection{Human Control of Robot Manipulators}
\textit{No intermediate representation--}
A simple baseline for teleoperation is the use of one or more static cameras, which the user can either see all at once, or switch between manually. While simple to implement, a static camera approach is limiting: they present the user with limited geometric information, and don't leverage computation to enhance human perception. Static cameras are also prone to being blocked by the robot manipulator itself. There are a number of approaches which improve on the static camera baseline without building an intermediate representation. Murata et. al. \cite{Murata2014} considers mobile manipulators (manipulators mounted on mobile robots), and renders a CAD model of the robot on top of a background made of stitched, observed images. The CAD model is kept updated to represent the robot's current state, and the user is able to position the camera to generate an arbitrary view. A different approach is taken by Rakita et. al. \cite{rakita2018autonomous}, in which a robot arm is used as a camera operator for another robot arm. The camera operator's pose is automatically optimized in real time to present a clear view of the manipulating robot's end effector.

\begin{figure}
    \centering
    \includegraphics[width=0.5\textwidth]{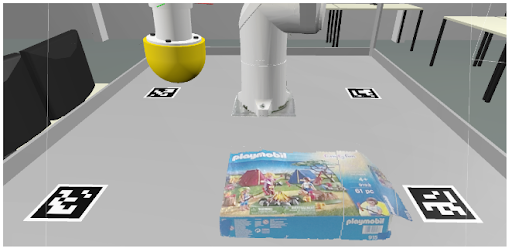}
    \caption{Kohn et. al. \cite{Kohn2018} uses meshes to represent known objects (table and robot) and point clouds to represent unknown objects (puzzle box) for teleoperation. Image from Kohn et. al.}
    \label{fig:Kohn}
\end{figure}

\textit{Point Clouds--} Point clouds are a common choice for representing scenes for real time operation, because they are the native output of depth cameras, and can be displayed in real time with little processing. Brizzi et. al. \cite{brizzi2017effects} considers augmented reality for VR teleoperation. Operators see an RGB image of the scene, along with features, such as distance to the target, direction to the target, or gripper state. Point clouds from a depth camera are used to calculate these features. Kohn et. al. \cite{Kohn2018} displays a combination of point clouds from RGBD cameras and meshes to a user in VR. Meshes are used for known objects (such as the floor, table, and robot) and point clouds are used for the unknown. The unknown objects are filtered in real time according to the meshes, as shown in \cref{fig:Kohn}. A similar approach is taken by Su et. al. \cite{su2019development}. Wei et. al. \cite{Wei2021} similarly displays point clouds to users, but unlike previous approaches, they do not use meshes to represent known objects, aside from the robot gripper. They perform a user study, comparing the point cloud to multi-view RGB and to a point cloud projected onto an RGB image. In their experiments, the hybrid point cloud and RGB view performs best.

\textit{Meshes--} While point clouds are native and fast to display, meshes may be preferred due to their potentially higher visual fidelity. Wonsick et. al. \cite{wonsick2021telemanipulation} takes an approach similar to the mesh-based approaches in \cref{sec:manipulation}; a point cloud of the scene is semantically segmented to identify its class amongst known objects. A known 3D model is then fit to each point cloud segment, and that 3D model is displayed to the user in virtual reality. Models are able to be constructed in much higher fidelity, and the scene can be entirely rendered, enabling control of lighting, texture, etc. They run a user study and say that users find their approach more usable and less cognitive load than point-cloud streaming alone. However their approach relies on having known 3D models of objects in the environment, and has no fallback if such a model does not exist, as such it is not suitable to unknown environments. Piyavichayanon et. al. \cite{piyavichayanon2022collision} generates a ``depth mesh'' by combining the view of multiple depth cameras. This mesh is used to display augmented reality features, such as distance to a collision state, on a handheld smartphone. Merwe et. al. \cite{merwe2019human} performs a user study to investigate how the type of information presented to the operator changes their performance. They compare a full information 3D model to a ``representative'' mesh based model, which only displays crucial information. They find that task completion time is lower with the full model, but cognitive load is higher. 


\textit{Occupancy Grids--}
Omarali et. al. \cite{omarali2020virtual} considers multiple modes of visualization for VR teleoperation. Alongside the depth camera-based baselines, they present a hybrid view, which displays the current output from the depth camera, alongside a translucent occupancy map in the previously observed areas. This occupancy map gives the user some context for the greater environment, while indicating that the context is not as reliable as the currently observed region. They find that users prefer the hybrid depth camera + occupancy map approach.


\subsection{Future Directions in Teleoperation}
\textbf{Improving real-time visualization of the robot's environment}. Existing approaches tend to rely on depth cameras, which can render the world in real time, but are low in detail and high in noise. Approaches which attempt to process this data into something more appealing, like a mesh, require heavy computation and require meshes of potential objects to be known a priori. Future work should consider ways to improve the fidelity and detail of reconstruction in real time and in unknown environments. To some extent, advances in imaging, such as high resolution depth sensors may improve this problem. Another potential solution is dynamically directing sensing towards areas that need high detail, such as around the end effector. The problem of representing the environment from a novel perspective is known as novel view synthesis and is well studied in the computer vision literature (see the recent review \cite{gao2022nerf}). Incorporating techniques from novel view synthesis could allow a more complete environment representation to be displayed to the operator.

\textbf{Building representations suited to the operator}. Currently, the vast majority of approaches focus on building representations which represent the environment faithfully. However, it has long been known that realistic displays are not always ideal, as they can make it difficult to parse what is relevant \cite{Smallman2005NaiveRM}. Some work covered in this review has considered building a sparse representation of only relevant features \cite{merwe2019human, wonsick2021telemanipulation}. However, these approaches require such a representation to be specified by hand prior to operation. Future work should consider ways of filtering the scene representation to represent only what is important to the user in an unknown scene.

\section{Future Directions}
\label{sec:future_research}
\begin{figure}
    \centering
    \includegraphics[width=\textwidth]{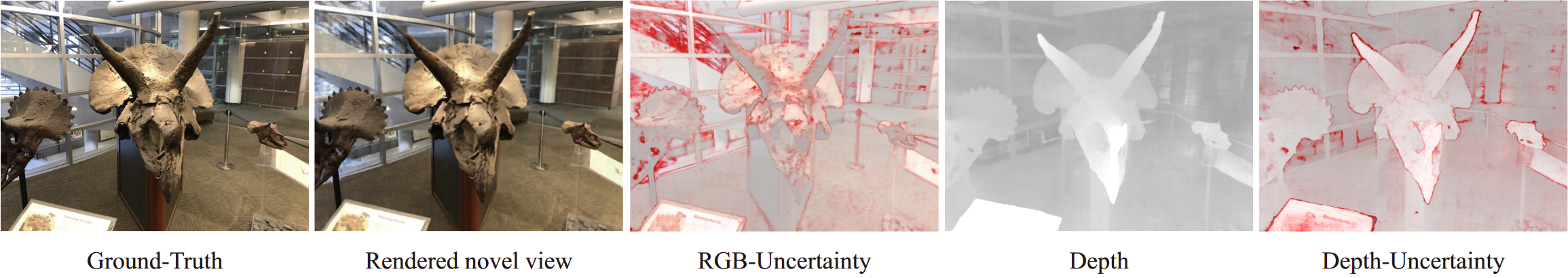}
    \caption{Shen et. al. \cite{shen2021stochastic} builds a scene representation which includes dense uncertainty estimates. A similar approach could prove useful to build scene representations for robot manipulators. Image from Shen et. al.}
    \label{fig:Shen2021}
\end{figure}

\textbf{Quantifying Uncertainty}. Present systems for robot manipulation build a \textit{most likely} map of the environment. In poor sensing conditions, this map may be highly unreliable, leading the downstream robotics application to make incorrect but fully confident choices based on the unreliable map. Future work should work to incorporate uncertainty into the scene representation itself, in order to enable robot policies which act intelligently in the face of uncertainty. When in a highly uncertain environment such a robot could, for example, ask a human for help, or gather additional information about its environment before taking action. The issue of uncertainty estimation is well studied in mobile robotics, especially in the context of SLAM \cite{rodriguez2018importance}. Recent approaches in novel view synthesis have also enabled dense estimates for geometric and photometric uncertainty \cite{shen2021stochastic, Sunderhauf2022}, as shown in \cref{fig:Shen2021}. Such approaches could be applied to robot manipulators, especially for problems like collision avoidance.

\textbf{Joint prediction of scene semantics and geometry}. Present approaches which infer scene semantics do so by taking some geometric representation of the scene as input, and producing a semantic map based on that geometry. With this approach, scene geometry informs scene semantics, but semantics has no influence on geometry. In reality, semantic information about an object can provide queues about its likely geometry. If an object is identified as a coffee mug, it probably has a handle on it. If an object is identified as a baseball, it is probably spherical. Floors and tabletops tend to be flat and level with one another. Future approaches could infer scene geometry and semantics jointly, by encoding sensor data in a learned latent representation, which is used to inform both geometric and semantic inference. Joint inference approaches have been used for semantic SLAM \cite{Bownman2017, sunderhauf2015slam, doherty2019multimodal}. Implementing such an approach for robot manipulators would enable higher fidelity predictive mapping and improved semantic understanding.


\textbf{Sensing-first development and real-world benchmarking}. Many existing approaches, particularly for grasping and collision avoidance, are benchmarked entirely in simulation. Real-world demonstrations are addressed after the development has been optimized for simulation, and often only occur under highly controlled conditions. Future work should consider developing methods with sensing in mind from the beginning, and aiming for high performance on real-world tasks with sensing, rather than in simulation. This means building systems which are robust to sensor noise, and able to act on raw sensor data in real time. One encouraging sign towards such a goal is the just announced RT-1 project from Google \cite{Brohan2022RT1}, which is an embodied agent that is benchmarked entirely on real-world tasks.

\textbf{Representations for Alternative Sensing Modalities}. All works addressed in this survey utilize data from typical vision based sensors: RGB cameras, depth cameras, and/or LiDaR. There are many other sensor types used in robotics, such as force torque sensors and tactile sensors, which provide information about the environment. However, existing approaches react immediately to information from these sensors, and do not build an intermediate representation which persists over time. Can we build scene representations which model the factors that these sensors measure? For example, the measurement from a tactile sensor may be influenced by how hard the object is, how heavy it is, or even how conductive it is, alongside the object's geometry. Present scene representations do not encode all of this information, making it very difficult to relate new observations from a tactile sensor to the known scene. Creating scene representations which do encode information from alternative sensors could enable applications such as SLAM or collision avoidance under low vision conditions.

\section{Conclusion}
Building new scene representations for robot manipulation is an important step towards creating fully autonomous embodied agents which can interact intelligently with the world. Continued advances in sensing and robotics will necessitate new representations which encode data from new sensors and support new robot form factors and applications. There are two challenges which are present in all works covered in this review: building representations which can be constructed in real time, and which are robust to sensor noise. Given the recent pace of advancement in robotics and computing, we are confident that these challenges will be sufficiently overcome.

\begin{table}[h!]
    \centering
    \begin{tabular}{p{0.3\linewidth}|p{0.3\linewidth}|p{0.3\linewidth}}
         \textbf{Application} &\textbf{Common \newline Representations} & \textbf{Less Common \newline Representations} \\
         \hline
         Collision Avoidance &  Signed distance fields & Convex hulls \newline Potential fields \newline Geometric primitives \newline Learned representations \\
         \hline
         Manipulation and Grasping & Direct action \newline Meshes \newline Point clouds & Voxel grids \newline Signed Distance Fields \newline Learned representations\\
         \hline
         Teleoperation & Point clouds & Meshes \newline Occupancy grids
    \end{tabular}
    \caption{Scene representations covered in this review}
    \label{tab:my_label}
\end{table}

\clearpage

\bibliography{sources}

\end{document}